\documentclass[preprint,12pt]{article}

\usepackage[margin=1in]{geometry}
\usepackage{amssymb}
\usepackage{amsmath}
\usepackage[]{algorithm2e}
\usepackage{csvsimple}
\usepackage{url}
\usepackage{multirow}
\usepackage{caption}
\usepackage{subcaption}
\usepackage[utf8]{inputenc}
\usepackage[english]{babel}
\usepackage{graphicx}
\usepackage{authblk}

\newcommand{\etal}{\textit{et al.}}
\DeclareMathOperator{\support}{support}
\DeclareMathOperator{\subpat}{sub}
\DeclareMathOperator{\parent}{parent}

\newtheorem{definition}{Definition}
\newtheorem{example}{Example}
\newtheorem{corollary}{Corollary}
\newtheorem{rmk}{Remark}

\begin{document}

\title{Extended Vertical Lists for Temporal Pattern Mining \\from Multivariate Time Series}

\date{}

\author[1]{\small Anton Kocheturov\thanks{Corresponding author. e-mail: antrubler@gmail.com}}
\author[2]{\small Petar Momcilovic}
\author[3]{\small Azra Bihorac}
\author[1]{\small Panos M. Pardalos}

\affil[1]{\footnotesize Center for Applied Optimization, Industrial \& Systems Engineering, University of Florida,
Gainesville, FL}
\affil[2]{\footnotesize Industrial \& Systems Engineering, University of Florida, Gainesville, FL}
\affil[3]{\footnotesize Division of Nephrology, Hypertension, \& Renal Transplantation, University of Florida, Gainesville, FL}

\maketitle

\begin{abstract}
Temporal Pattern Mining (TPM) is the problem of mining predictive complex
temporal patterns from multivariate time series in a supervised setting.
We develop a new method called the Fast Temporal Pattern Mining with Extended Vertical Lists.
This method utilizes an extension of the Apriori property which requires a more complex pattern
to appear within records only at places where all of its subpatterns are detected as well.
The approach is based on a novel data structure called the Extended Vertical List that tracks
positions of the first state of the pattern inside records.
Extensive computational results indicate that the new method performs
significantly faster than the previous version of the algorithm for TMP.
However, the speed-up comes at the expense of memory usage.
\end{abstract}

\section{Introduction}
\label{intro}

Continuously expanding resources for computing, data storage,
and transmission have been enabling analyses of complex data sets 
emerging from various domains such as medical event detection and prediction 
\cite{han2001prefixspan,ayres2002sequential,wang2004bide,chiu2004efficient},
fraud detection \cite{iyer2011credit}, \textit{etc}.
We consider the problem of extracting temporal patterns
 from multivariate time series records in a supervised setting.
Our main contribution is a faster algorithm for mining 
class-specific patterns having temporal relations between their states.
Our motivation is to develop an algorithm that can
 be built into a workflow for real-time analytic engines.
The framework for frequent pattern mining was 
developed in \cite{agrawal1995mining,srikant1996mining}.
Several extensions have been introduced 
\cite{han2001prefixspan,ayres2002sequential,wang2004bide,chiu2004efficient}
which were successfully utilized in medical decision-making
\cite{sacchi2007data,hauskrecht2013data,batal2016efficient,moskovitch2015classification}.
Data  is collected in the form of a set of records 
where each record is characterized by numerical
(e.g., heart rate or blood pressure) and categorical 
(e.g., an indicator of whether a patient is on medication or not)
time series combined with categorical and numerical attributes like gender, age, \textit{etc}.

In this paper, the work of Batal \textit{et al.} \cite{batal2016efficient} is extended by
introducing a new algorithm called the Fast Temporal Pattern Mining with Extended Vertical Lists. 
The idea is to utilize the Apriori \cite{agrawal1994fast} property on the level of pattern positions inside records.
%Without introducing precise definitions of Temporal Pattern (TP) and
%other related concepts as subpatterns and others (see Section \ref{definitions}),
Informally, we require the Apriori property to state that a Temporal Pattern (TP)
may appear only in the records where all of its subpatterns appear as well.
For example, if TP ``Heart Rate is Very High \textit{before} Blood Pressure is Low" is found in record $i$, then both its subpatterns
``Heart Rate is Very high" and ``Blood Pressure is Low" must appear in record $i$.
In this form, the Apriory property was used to reduce the search space
for mining TPs \cite{batal2016efficient}
and similar notions as itemsets \cite{zaki2000scalable} and sequential patterns \cite{ayres2002sequential,zaki2001spade}
via the vertical data format \cite{zaki2000scalable}.
We suggest a new data structure called the Extended Vertical List that keeps track of positions of the first state of the TP inside the records and links them to the positions of the pattern parent (a subpattern without the first state) inside the records.
This idea allows us to reduce the computational time by a factor of several hundreds
on a number of datasets (see Section \ref{computationalresults} for details).
The speed-up comes at the cost of increased memory consumption, which is a typical trade-off in such a kind of algorithms.

\section{Concepts and Definitions}
\label{definitions}

Assume dataset $D$ of $n$ records $d_i,~i=1,\dots,n$, where each record is composed of
$m$ time series $x_j^i \in X_j$ and an outcome, or class label, $y_i \in Y$ associated with it:
$d_i = \left( x_1^i,x_2^i,\dots,x_m^i,y_i \right).$
The overall goal is to train classifier $f:X_1 \times \dots \times X_m \rightarrow Y$,
\textit{i.e.}, we consider the Supervised Multiple Time Series Classification Problem.
In such problems, time series are typically mapped first into a feature
domain with feature extraction techniques:
$g:X_1 \times \dots \times X_m \rightarrow F_1 \times \dots \times F_k$.
In this paper, we focus on improving computational aspects of function $g$.
In this section, we follow the definitions given in \cite{batal2016efficient}
 with slight modifications for the presentation to be self-contained.

We start with reducing dimensionality through converting each time series into a set of temporal abstractions
in the form $$\left<(V_1,s_1,e_1),\dots,(V_k,s_k,e_k)\right>,$$ where $V_i \in \Sigma$
is a temporal abstraction that is in effect from start time $s_i$ till end time $e_i$,
e.g., temporal abstraction (``Low'',5,12) means that the time series was low from time 5 till time 12;
$\Sigma$ is the alphabet, or set of possible abstractions (e.g., $\Sigma = \{\text{``Low'', ``Normal'', ``High''}\}$).
For a given set of temporal abstractions, we also require $s_1 \leq e_1 \leq s_2 \dots \leq s_k \leq e_k$,
meaning that no abstraction can start earlier than any previous one finishes.
%Based on common logic, we also forbid two consecutive temporal abstractions
%inside the same time variable to be represented by a single time stamp
%(e.g, $s_k = e_k = s_{k+1} = e_{k+1}$). 
%Those are rather technical constraints but they are important for constructing temporal patterns.

The alphabet $\Sigma$ can be defined in several ways. In this paper, we focus
 on value and trend abstractions; particular examples include:

\begin{enumerate}
\item \textbf{Value Abstractions:}
$\Sigma = \{\downarrow \downarrow, \downarrow, \textendash, \uparrow, \uparrow\uparrow \}$ where $\downarrow \downarrow$, $\downarrow$, $\textendash$, $\uparrow$, and $\uparrow\uparrow$ stand for ``Very Low'', ``Low'', ``Normal'', ``High'' and ``Very High'', respectfully.
Exact ranges for transformation may be set up by a field expert.
In our computations, we used time-series-specific percentiles $\{0.1, 0.25, 0.75, 0.9\}$.
\item \textbf{Trend Abstractions.}
$\Sigma = \{\rightarrow,\nearrow,\searrow \}$ where $\rightarrow$, $\nearrow$, and $\searrow$ stand for ``Steady'', 
``Increasing'', and ``Decreasing'', respectfully.
For this transformation, we used the approach by \cite{Keogh2004segmenting}.
\end{enumerate}

%If one decides to combine several ways and let the time abstractions overlap, copying the time series and applying one way per copy will solve the issue.

Now we are ready to define the Multivariate State Sequence which 
is a representation of the corresponding multivariate time series in alphabet $\Sigma$.

$ $
\begin{definition}
$ $
\begin{itemize}
\item
$S = (F,V)$ is a \textbf{state} where is $F$ is a variable label and $V \in \Sigma$ is an abstraction  value.
\item
$E = (F,V,s,e)$ is a \textbf{state interval} where $(F,V)$ is a state
and $s$ and $e$ are the start and end times of the state interval.
\item
$Z = \langle E_1,\dots,E_l \rangle$ is a \textbf{Multivariate State Sequence (MSS)}
with the states sorted according to the non-decreasing order of their start times:
$E_i.s \leq E_{i+1}.s, 1 \leq i \leq l-1$.
\end{itemize}
\end{definition}

\begin{figure}[h]
\centering
\includegraphics[width=0.8\textwidth]{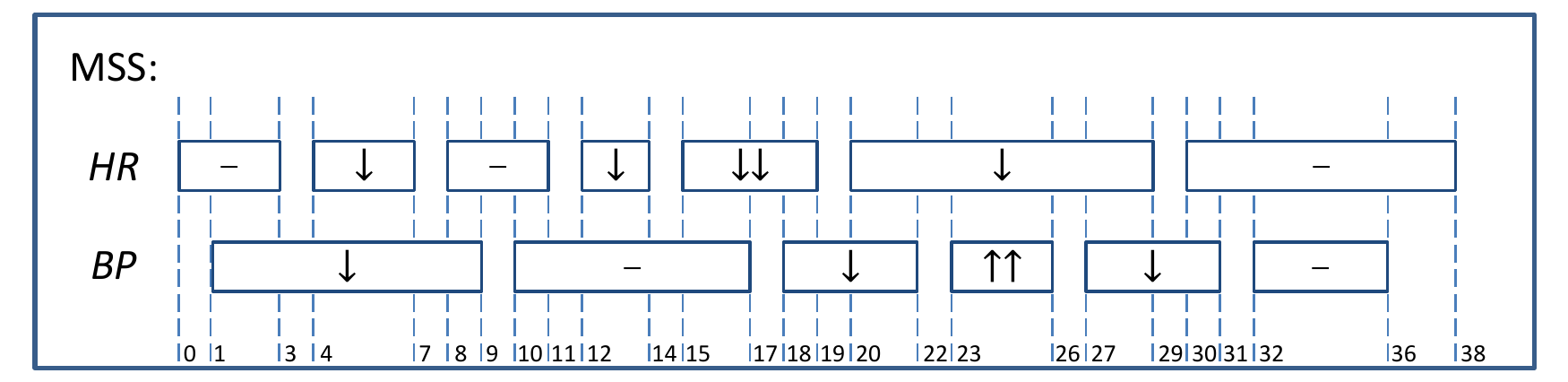}
\caption{An example of an MSS with time variables Heart Rate (HR) and Blood Pressure (BP).}
\label{fig:MSSExample}
\end{figure}

\begin{example}
\footnotesize
$S = (\text{HR}, \downarrow)$ is a \textit{state} indicating that temporal variable Heart Rate is on a low level,
while \textit{state interval} $E = (\text{HR},\downarrow,12,15)$ extends the state by including information on its start and end time moments.
Finally, an MSS combines several state intervals coming from different time series as in
MSS $Z = \langle E_1 = (\text{HR},\textendash,0,3)$,
$E_2 = (\text{BP},\downarrow,1,9)$,
$E_3 = (\text{HR},\downarrow,4,7)$,
$E_4 = (\text{HR},\textendash,8,11)$,
$E_5 = (\text{BP},\textendash,10,17)$,
$E_6 = (\text{HR},\downarrow,12,14)$,
$E_7 = (\text{HR},\downarrow\downarrow,15,19)$,
$E_8 = (\text{BP},\downarrow,18,22)$,
$E_9 = (\text{HR},\downarrow,20,29)$,
$E_{10} = (\text{BP},\uparrow\uparrow,23,26)$,
$E_{11} = (\text{BP},\downarrow,27,31)$,
$E_{12} = (\text{HR},\textendash,30,38)$,
$E_{13} = (\text{BP},\textendash,32,36)\rangle$
(see Figure\ref{fig:MSSExample}). 
\end{example}

Temporal Pattern is the next level of abstraction, which allows to  remove 
exact values of start and end times
and focus on temporal relationships of the state intervals. For this purpose,
Allen's logic is often used \cite{allen1984towards}, where only two time relations out of 13 possible 
are chosen (see \cite{batal2016efficient} for a relevant discussion).

For two state intervals $E_i$ and $E_j$ with $E_i.s \leq E_j.s$, we say that $E_i$ finishes \textbf{before} $E_j$ 
if $E_i.e < E_j.s$ and denote it as $R(E_i,E_j) = b$.
Otherwise, we say that $E_i$ \textbf{co-occurs} with $E_j$ and denote it as $R(E_i, E_j) = c$.

\begin{definition}
$ $
\label{def:TP}
\begin{itemize}
\item
$P = \left( \langle S_1,\dots,S_k\rangle,R \right)$ is a \textbf{Temporal Pattern} (TP) of size $k$ $(|P| = k)$ with states $S_1,\dots,S_k$,
where $R$ is a (upper-triangular) matrix describing pair-wise temporal relationships between the states:
$R_{i,j} \in \{b,c\},~1 \leq i < j \leq k$. \footnote{$R_{i,j}$ is \textbf{defined} for \textit{states} $i$ and $j$ of the pattern, while $R(E_i,E_j)$ is \textbf{computed} for \textit{state intervals} $i$ and $j$ of the MSS.}

\item
Given MSS $Z = \left<E_1, E_2,\dots, E_l \right>$
and temporal pattern $P = (\left< S_1,\dots, S_k \right>, R)$ $(k \leq l)$,
we say that $Z$ \textbf{contains} $P$, denoted as $P \in Z$, 
if there is a mapping $\pi : \{1,\dots,k\} \to \{1,\dots,l\}$ that matches every state $S_i$
 in $P$ to a state interval $E_{\pi(i)}$ in $Z$ such that:

\begin{enumerate}
\item
$S_i.F = E_{\pi(i)}.F~\text{and}~ S_i.V = E_{\pi(i)}.V,~1 \leq i \leq k$,
\item
$\pi(i) < \pi(j),~i < j$,
\item
$R(E_{\pi(i)},E_{\pi(j)}) = R_{i,j},~i < j$.
\end{enumerate}

\end{itemize}

\end{definition}

The first requirement guarantees that each state of $P$ matches an appropriate state interval in $Z$,
 while the rest of the constrains enforce that the temporal relations in  $P$ correspond to a correct overlapping
 of the state intervals in $Z$.

\begin{figure}
\centering
\begin{subfigure}{.24\textwidth}
\centering
\includegraphics[width=0.8\textwidth]{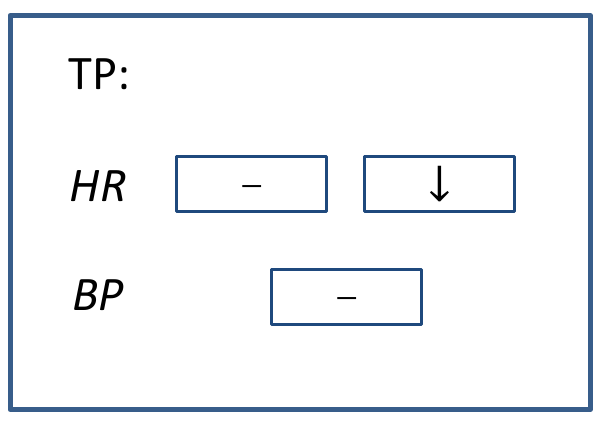}
\caption{}
\label{fig:TPExample}
\includegraphics[width=0.8\textwidth]{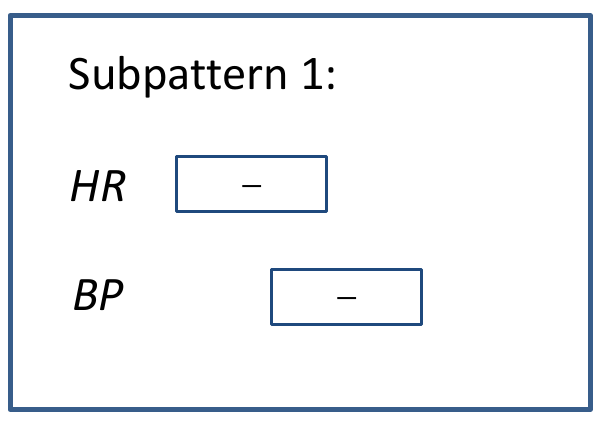}
\caption{}
\label{fig:SubTPExample}
\end{subfigure}%
\begin{subfigure}{.24\textwidth}
\centering
\includegraphics[width=0.8\textwidth]{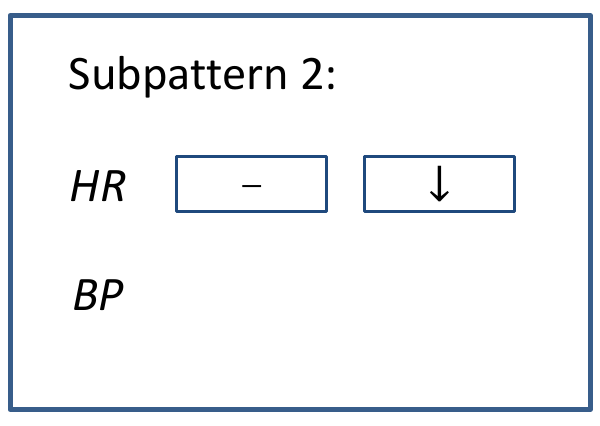}
\caption{}
\label{fig:Sub2TPExample}
\includegraphics[width=0.8\textwidth]{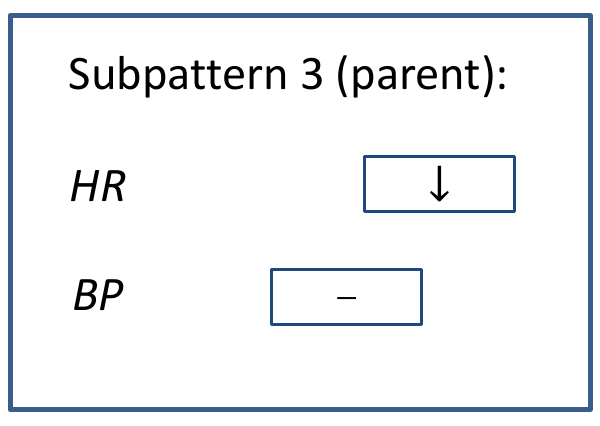}
\caption{}
\label{fig:Sub3TPExample}
\end{subfigure}
\caption{TP and its subpatterns: (a) an example of a TP; (b) the subpattern without the last state; (c) the subpattern without the middle state; (d) the parent, the subpattern without the first state. }
\label{fig:TPandSubs}
\end{figure}

\begin{example}
\footnotesize
$P = \left( \langle S_1,S_2,S_3\rangle,R \right)$ is a TP of size $3$ (see Figure \ref{fig:TPExample})
 with states $S_1 =(\text{HR},\textendash)$,
$S_2 =(\text{BP},\textendash)$,
$S_3 =(\text{HR},\downarrow)$ and relationships
matrix $R = \left( R_{1,2},R_{1,3},R_{2,3}\right)$, where
$R_{1,2} = c$, $R_{1,3} = b$, and $R_{2,3} = c$.
The MSS from Figure \ref{fig:MSSExample} contains this TP since
MSS's state intervals $E_4 = (\text{HR},\textendash,8,11)$,
$E_5 = (\text{BP},\textendash,10,17)$, and
$E_6 = (\text{HR},\downarrow,12,14)$
match the states of $P$ and the time relationships
are satisfied: for example, $E_4.e = 11 > 10 = E_5.s$
and, therefore, state intervals 4 and 5 co-occur, or $R(E_4,E_5) = c$,
and equals the specified time relationship $R_{1,2}$ between states 1 and 2 of $P$.   
\end{example}

\begin{definition} $\widetilde{P}$, $|\widetilde{P}| = \widetilde{k}$, is a \textbf{subpattern} of $P$, $|P| = k$ ($\widetilde{k} \leq k$),
denoted as $\widetilde{P} \subset P$, if there is a mapping $\pi:  \{1,\dots,\widetilde{k}\} \to \{1,\dots,k\}$ such that:

\begin{enumerate}
\item
$\widetilde{S}_i = S_{\pi(i)}, 1 \leq i \leq \widetilde{k}$, where $\widetilde{S}_i$ and $S_{\pi(i)}$  are states in $\widetilde{P}$ and $P$, respectfully,
\item
$\pi(i) < \pi(j), i < j$,
\item
$\widetilde{R}_{i,j} = R_{\pi(i),\pi(j)}, 1 \leq i < j \leq \widetilde{k}$.
\end{enumerate}
\end{definition}

\begin{example}
\footnotesize
Temporal pattern $P$ from Figure \ref{fig:TPExample} has three subpatterns:
the subpattern without the last state (Figure \ref{fig:SubTPExample}),
the subpattern without the middle state (Figure \ref{fig:Sub2TPExample}),
and its parent, denoted as $parent(P)$, the subpattern without the first state (Figure \ref{fig:Sub3TPExample}).
\end{example}

It is straightforward to show:

\begin{corollary}
\label{cor:Apriori}
If $\widetilde{P} \subset P$ and $P \in Z$, then $\widetilde{P} \in Z$.
\end{corollary}

This corollary, also known as the Apriori property, brings the main computational improvement to the framework.

The overall goal is to mine class-specific temporal patterns which appear in a number of MSSs belonging to a certain class.
For this purpose, we use the threshold $\theta  \in [0,1]$ and define the minimum support.
Assume that $D = \{Z_1,\dots,Z_n\}$ is a data set of $n$ MSSs
and $Y=\{y_1,\dots, y_c\}$ is a set of possible classes, or outcomes.
Let $D_i$ denote a set of records from $D$ which belong to class $y_i$
(each record belongs to exactly one class).
$Z_j \in D_i$ denotes that record $j$ is in class $y_i$.

\begin{definition}
\label{def:FTP}
$ $
\begin{itemize}
\item
For a given  temporal pattern $P$ and class $y$, we define \textbf{support} of $P$ in class $y$,
denoted as $\support(P, D_y)$, as a number of MSSs from $D_y$ that contain $P$:

\begin{equation*}
\support(P, D_y) = |\{Z \in D_y:~P \in Z\}|.
\end{equation*}

\item
$P$ is a \textbf{Frequent Temporal Pattern (FTP)} in $D$, if for some class $y$
\label{def:FTP}
\begin{equation*}
\support(P, D_{y}) \geq  \theta \times |D_{y}|.
\end{equation*}
\end{itemize}
\end{definition}

In other words, $P$ is an FTP in $D$ if the proportion of MSSs 
containing $P$ is not smaller than threshold $\theta$ for at least one class.

\begin{corollary}
\label{cor2}
If $\widetilde{P} \subset P$ and $\widetilde{P} \text{ is \textbf{not} FTP in D}$, then $P$ is \textbf{not} FTP in D.  
\end{corollary}

Corollary \ref{cor2} is a straightforward consequence of  Corollary \ref{cor:Apriori} and Definition \ref{def:FTP}.

%\section{Mining Frequent Temporal Patterns}
%\label{ftpm}

The Frequent Temporal Pattern Mining algorithm (FTPM), as it was given in \cite{batal2016efficient,batal2009multivariate,batal2011pattern,batal2012mining}, is
a breadth-search procedure for finding all FTPs. First, all FTPs of size 1 are found.
Then a list of candidate TPs of size 2 is generated. After that, each candidate TP is validated
for being an FTP and a list of FTPs of size 2 is formed. The procedure is repeated until all FTPs are found
 or some stopping criteria are met, e.g., size is no more than a predefined value $k_{max}$
(see Algorithm \ref{algorithm1}).
Other schemes like depth-first search are possible, but the breadth-search paradigm is important for
eliminating incoherent candidate TPs as it can be be seen later.

\begin{algorithm}
\footnotesize
 \KwIn{\textit{D, FTPs of size 1}}
 \KwOut{\textit{FTPs} }
 \textit{FTPs} $\gets$ \textit{FTPs of size 1}\\
 \textit{new-FTPs} $\gets \textit{FTPs of size 1}$ \\
 \While{$|\text{new-FTPs}| > 0$ \textbf{and no} \textit{other criteria are met}}{
  \textit{candidates} $\gets$ \textit{CreateCandidates$($new-FTPs, FTPs of size 1$)$} \\
  \textit{new-FTPs} $\gets \emptyset$ \\
  \ForAll{$P \in$ candidates}{
   \If{$P$ is FTP in D}{
    \textit{new-FTPs $\gets$ new-FTPs $\cup \{P\}$} \\
   }
  }
  \textit{FTPs $\gets$ FTPs $\cup $ new-FTPs} \\
 }
 \caption{FTPM algorithm: the general framework.}
 \label{algorithm1}
\end{algorithm}

The most computationally expensive part of this framework is validating if a candidate TP is frequent.
Thus, further careful elimination of infrequent TPs at the step of creating candidates is very important.
Based on Corollary \ref{cor2}, a TP is frequent only if all of its sub-patterns  are frequent.
for a pattern of size $k$, we need to verify only if subpatterns of size $k-1$ are frequent due to transitivity.

An idea of assigning to each FTP a list of record identifiers 
which contain it: $P.ids = \{i: P \in Z_i\}$,
reduces the search space drastically \cite{batal2016efficient}.
It is based on the vertical data format \cite{zaki2001spade,zaki2000scalable}.  Due to Corollary \ref{cor:Apriori},
a candidate TP of size $k+1$ will appear only in records where all its subpatterns appear as well.
Therefore, we need to check only its $k$-subpatterns because record id lists of the subpatterns of
smaller sizes includes the list for at least one $k$-subpattern (for which it is its subpattern).   
Such a list is called the list of \textbf{potential records}:

\begin{equation*}
P.p\_ids = \mathop{\cap}_{\widetilde{P} \in \subpat(P)}{\widetilde{P}.ids} = \mathop{\cap}_{\widetilde{P} \in \subpat_k(P)}{\widetilde{P}.ids},
\end{equation*}
where $\subpat(P) = \{\widetilde{P}: \widetilde{P} \subset P\}$ and $\subpat_k(P) = \{\widetilde{P}: \widetilde{P} \subset P~\text{and}~|\widetilde{P}| = k\}$.

If for all classes number of the potential records is smaller
than the corresponding minimal support values, then this pattern is not frequent, and it can be discarded.

\section{Frequent Temporal Pattern Mining with Extended Vertical Lists}
\label{ftpmwel}

In this section, we present our approach for Frequent Temporal Pattern Mining. The main idea is that,
for given MSS and FTP, we keep track of positions (indices of the state intervals in the MSS) where the first state of
the pattern appears inside the record. We say that the pattern starts at those positions.

Assume that FTPs of all sizes $1,\dots,k$ have been found.
A  coherent candidate temporal pattern $P$ ($|P|=k+1$) constructed from FTP $P_0$ ($|P_0|=k$) and state $S$
(see \cite{batal2016efficient} for relevant discussion),
has exactly $k+1$ subpatterns of size $k$ ($|\subpat_k(P)| = k+1$). 
Some subpatterns may be identical: for example, all subpatterns of size 2 are the same for the pattern
with 3 identical sates $(\text{HR},\textendash)$,$(\text{HR},\textendash)$, and $(\text{HR},\textendash)$.
From $\subpat_k(P)$, no more than $k$ patterns (some may be identical) start with state $S$ and all of them are in $$\subpat_k(P) \setminus \parent(P).$$
It is straightforward to see that $P$ cannot start at a position $i$ inside $z$ if at least one of the subpatterns
from $\subpat_k(P) \setminus \parent(P)$ does not start at the same position.

\begin{example}
\label{ex4}
\footnotesize
Assume that we want to find if MSS $z$ (Figure \ref{fig:MSSExample}) contains
temporal pattern $P$ (Figure \ref{fig:TPExample}):
$$P = (\left< (\text{HR},\textendash),(\text{BP},\textendash),(\text{HR},\downarrow) \right>, R)$$
with $R_{1,2} = c,R_{1,3} = b,R_{2,3} = c$.
Pattern $P$ has two subpatterns $P_{1}$ (Figure \ref{fig:SubTPExample}) 
and $P_{2}$ (Figure \ref{fig:Sub2TPExample})  which have the same first state $(\text{HR},\textendash)$:
$$P_{1} = (\left< (\text{HR},\textendash),(\text{BP},\textendash)\right>, R_{1,2} = c),$$
$$P_{2} = (\left< (\text{HR},\textendash),(\text{HR},\downarrow)\right>, R_{1,2} = b).$$
$P_{1}$ starts at positions 4 and 12 in $z$, while $P_{2}$ starts at positions 1 and 4.
Thus, $P$ \textit{may potentially} start only at position 4 where both the subpatterns start.
Those positions are potential because there are also time relationships between the states which were not checked yet.
\end{example}

\begin{rmk}
\label{remark1}
\footnotesize
$P_2$ appears 5 times in MSS $z$ because states $(\text{HR},\textendash)$ and $(\text{HR},\downarrow)$
of  $P_2$ match the following pairs of the state intervals of $Z$: $(E_1,E_3)$,$(E_1,E_6)$,$(E_1,E_9)$,
$(E_4,E_6)$, and $(E_4,E_9)$ (in all the cases, time relationship $R_{1,2}=b$ is satisfied).
But we make the positions of the first state be relevant only, therefore, positions 1 and 4 are used.
\end{rmk}

One may want to store all possible appearances of $P$ in MSS $z$,
but the number of such appearances may grow rapidly: see, for example, Remark \ref{remark1}.
This requires significant memory storage. In turn, storing only starting positions of $P$ in the MSS
requires significantly less memory since the starting  
positions are always inside the intersection of the starting positions of the subpatterns form
$\subpat_k(P) \setminus \parent(P)$. Therefore, the number of starting positions is a non-increasing
function of pattern size. Such a trade-off gives a desired speed-up under a reasonable memory consumption increase:
see Section \ref{computationalresults}.
 
In general, for each TP we assign an Extended Vertical List (EVL),
a structure containing information on which MSSs contain the TP, starting positions
of the TP inside the MSSs, and the indices of (or links to) the starting 
positions of the \textit{parent} of the TP inside the MSSs.

\begin{definition}
$ $
\begin{itemize}
\item
Let $P.EVL$ denote \textbf{Extended Vertical List} associated with $P$.
\item
Let $P.EVL[z].pos$ denote \textbf{starting positions} of $P$
(positions of the first state of $P$) inside MSS $z$.
\item
Let $P.EVL[z].ind$ denote \textbf{indices} of \textit{specific} starting positions of $parent(P)$
(positions of the first state of $parent(P)$) inside MSS $z$.
For position $i \in  P.EVL[z].pos$, a corresponding specific index of the parent position is the
index of the smallest parent position in $z$ which is larger than $i$:
$$P.EVL[z].ind[i] = \min{\{j: \widetilde{P}.EVL[z].pos[j] > p\}},$$
where $\widetilde{P} = parent(P)$ and $p = P.EVL[z].pos[i]$.
\end{itemize}
\end{definition}
%
%Thus, $P.ppos[i]$ can be computed as the following:
%$$P.p\_pos[i] = \mathop{\cap}_{\widetilde{P} \in X} {\widetilde{P}.pos[i]},$$
%where $X = \subpat_k(P) \setminus \parent(P)$.
%

\begin{example}
\footnotesize
TP $P = \left( \langle S_1,S_2,S_3\rangle,R \right)$ (see Figure \ref{fig:TPExample})
 has states $S_1 =(\text{HR},\textendash)$,
$S_2 =(\text{BP},\textendash)$,
$S_3 =(\text{HR},\downarrow)$ and relationships
matrix $R = \left( R_{1,2},R_{1,3},R_{2,3}\right)$, where
$R_{1,2} = c$, $R_{1,3} = b$, and $R_{2,3} = c$.
Its parent is $P_0 = parent(P) = \left( \langle (\text{BP},\textendash), (\text{HR},\downarrow) \rangle, (R_{1,2} = c) \right)$
(see Figure \ref{fig:Sub3TPExample}). 
In turn, the parent of $P_0$ consists of a single state:
$P_{00} = parent(P_0) = parent(parent(P)) =  \left( \langle (\text{HR},\downarrow) \rangle, \emptyset \right)$.

Now, for MSS $z$ from Figure \ref{fig:MSSExample},
$P_{00}.EVL[z].pos = \{3,6,9\}$ because state $(\text{HR},\downarrow)$ corresponds to the state intervals
$E_3, E_6$, and $E_9$ of $z$. $P_{00}.EVL[z].ind = \emptyset$ since $P_{00}$ does not have a parent.
$P_{0}.EVL[z].pos = \{5\}$ because state $(\text{BP},\textendash)$ corresponds to the state interval
$E_5$ of $z$ (see Remark \ref{remark1}). $P_{0}.EVL[z].ind = \{2\}$.
Finally, $P.EVL[z].pos = \{4\}$ and $P.EVL[z].ind = \{1\}$.
\end{example}

Now we are ready to present the pseudo-code of
the Fast Temporal Pattern Mining with Extended Vertical Lists (FTPMwEVL) algorithm: see Algorithm \ref{alg:ftpmwel}.

\begin{algorithm}[h]
\footnotesize
 \KwIn{\textit{D, FTPs of size 1}}
 \KwOut{\textit{FTPs} }
 \textit{FTPs} $\gets$ \textit{FTPs of size 1}\\
 \textit{new-FTPs} $\gets \textit{FTPs of size 1}$ \\
 \While{$|\text{new-FTPs}| > 0$ \textbf{and no} \textit{other criteria are met}}{
  \textit{candidates} $\gets$ \textit{CreateCandidates$($new-FTPs, FTPs of size 1$)$}\\ 
  \textit{new-FTPs} $\gets \emptyset$ \\
  \ForAll{$P \in$ candidates}{
   $exposure \gets exposure(P)$ \\
   \lIf{\textbf{not} FindPotentialPositionsAndIndices(D,P)} { continue}

	\ForAll{id $\in P.p\_ids$}{
 	$new\_positions \gets \emptyset$ \\
 	$new\_indices \gets \emptyset$\\
 	$i \gets 1$\\
	\While{$i \leq |P.EVL[id].pos|$}{
		$pos \gets P.EVL[id].pos[i]$\\
		$ind \gets P.EVL[id].ind[i]$ \\
		
		$index \gets Search(parent(P), D, id, ind, \{pos\}, exposure)$\\
		
		\If{$index > -1$}{
			$new\_positions \gets new\_positions \cup \{pos\}$\\
			$new\_indices \gets new\_indices \cup \{index\}$
		}
		$i \gets i + 1$\\
 	}
 	$P.EVL[id].pos \gets new\_positions$\\
 	$P.EVL[id].ind \gets new\_indices$\\
	\lIf{$P.EVL[id].pos = \emptyset$}{$P.p\_ids \gets P.p\_ids \setminus id$}
	}
	\If{P is FTP in D}{
    		\textit{new-FTPs $\gets$ new-FTPs $\cup \{P\}$}
   	}
    }
  \textit{FTPs $\gets$ FTPs $\cup$ new-FTPs}
 }
 \caption{The Fast Temporal Pattern Mining with Extended Vertical Lists algorithm.}
 \label{alg:ftpmwel}
\end{algorithm}

The EVL data structure allows to achieve three main results.
First, it reduces the number of potential starting positions of a candidate TP $P$ by intersecting
the starting positions of its subpatterns from $\subpat_k(P)\setminus\parent(P)$ as in Example \ref{ex4}
and later linking the potential positions to the smallest starting positions of $parent(P)$.
Therefore, EVL reduces the number of candidate TPs to check in general
(see Algorithm \ref{alg3} for the pseudo-code).
For some MSSs from the dataset, the set of potential starting positions may be empty after the intersection
meaning that these MSSs will never contain $P$ and they can be skipped.

\begin{algorithm}[h]
\footnotesize
 \KwIn{\textit{D, P}}
 \KwOut{Boolean}
 
\textit{subpatterns} $\gets \subpat_k(P) \setminus \parent(P)$ \;
 
\lIf{\textit{subpatterns} $=\emptyset$} { \Return False}
$P.p\_ids = \mathop{\cap}_{\widetilde{P} \in \subpat_k(P)}{\widetilde{P}.ids}$\;
 
\lIf{\textbf{not} PotentiallyFrequent$(P)$}{ \Return False}

\ForAll{id $\in P.p\_ids$}{
 $P.EVL[id].pos \gets  \mathop{\cap}_{\widetilde{P} \in \textit{subpatterns}}{\widetilde{P}.EVL[id].pos}$ \;
 	$i = 1$\;
	\While{$i \leq |P.EVL[id].pos|$}{
		$pos = P.EVL[id].pos[i]$ \;
	
		\eIf{$\{j : parent(P).EVL[id].pos[j] > pos\} = \emptyset$}{
			$P.EVL[id].pos \gets P.EVL[id].pos \setminus pos$\;
		}
		{
			$P.EVL[id].ind[i] = \min{\{j : parent(P).EVL[id].pos[j] > pos\}}$ \;
			$i \gets i +1$ \;
		}
 	}
 \lIf{$P.EVL[id].pos = \emptyset$}{$P.p\_ids \gets P.p\_ids \setminus id$}
}

\eIf{\textbf{not} PotentiallyFrequent$(P)$}{ \Return False;} {\Return True;}

\caption{Function $FindPotentialPositionsAndIndices(D,P)$.}
\label{alg3}
\end{algorithm}

Second, to verify that a candidate TP $P$ is indeed inside FTP $z$,
we need to check that the the states of $P$ match the state intervals of $z$ and
the temporal relationships are satisfied according to Definition \ref{def:TP}. However, instead of looking
 through all possible combinations of appropriate state intervals,
EVL allows to check a significantly smaller amount of state intervals combinations:
we need to check only the  possible starting locations of $P$,
from which we can navigate directly to the appropriate state intervals matching the first state of $parent(P)$.
But these are the already found starting positions of $parent(P)$, therefore, we may skip some state intervals
matching  the first state of $parent(P)$.
Then we navigate directly to $parent(parent(P))$, and so on
(see Algorithm \ref{alg4}).

\begin{figure}[h]
\centering
\includegraphics[width=0.48\textwidth]{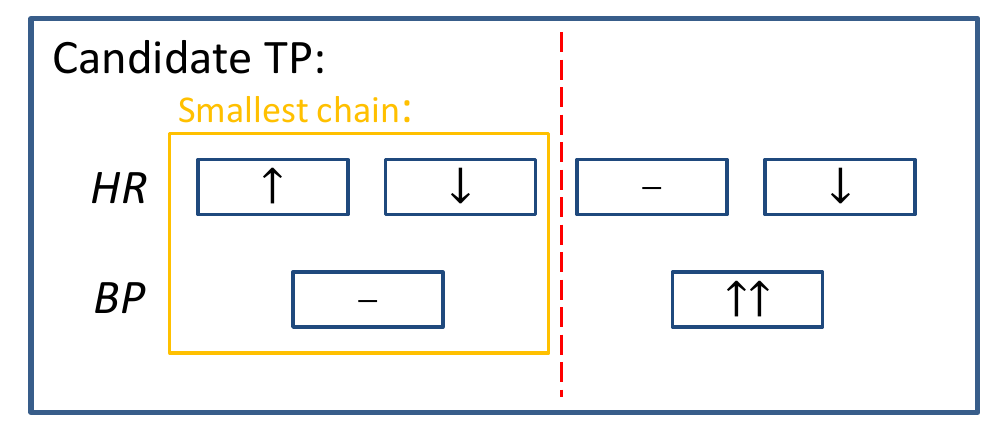}
\caption{An example of an TP and its smallest starting chain.}
\label{fig:chain}
\end{figure}

Third, EVL allows to check only a portion of the states of $P$.
For this purpose, the find \textbf{the smallest starting chain} of $P$.
By the smallest starting chain we mean a non-empty subpattern $P_{chain}$ at the beginning of $P$,
such that all states of $P_{chain}$ are strictly \textit{before} the remaining states of $P$
(see Figure \ref{fig:chain} for an example). 
For many long patterns the corresponding
smallest starting chain may be relatively small. 
Let us denote a subpattern of $P$ formed by
the remaining states of $P$ as $P_{end}$. 

When we check if $P$ is inside an MSS,
we need to traverse only the states of $P_{chain}$ and the first state of $P_{end}$
if any is present because $P_{chain}$ may be pattern $P$ itself. It is easy to see since
after we have arrived at $P_{chain}$ (we know it is there at this starting position)
 by recursive search function (Algorithm \ref{alg4}) and
checked that all time relationship between the states of $P_{chain}$ are satisfied, we need to verify
only that all the states of $P_{chain}$ are \textit{before} the first state of $P_{end}$ since all the states
of $P_{chain}$ will be before all the other states of $P_{end}$ by transitivity.
Thus, we need to check  $|P_{chain}| + 1$ first  states of $P$, if $P \neq P_{chain}$,
and $|P_{chain}|$ first states of $P$, otherwise. We call this number \textit{exposure} of $P$ and
denote it as $exposure(P)$.

\begin{algorithm}[h]
\footnotesize
 \KwIn{\textit{P, D, id, i, positions, exposure }}
 \KwOut{Integer}
 
\While{$i \leq |P.EVL[id].pos|$}{
	\If{Check accumulated time relationships in D[id]}{
		\lIf{$exposure = 0$}{\Return $~i$}
		$pos \gets P.EVL[id].pos[i]$\\
		$ind \gets P.EVL[id].ind[i]$ \\
		
		\Return $Search(parent(P), D, id, ind, positions \cap \{pos\}, exposure -1)$ \\
		}
	$i \gets i +1$ \\
}

\Return $~-1$
 
\caption{Function $Search(P, D, id, ind, positions, exposure)$.}
\label{alg4}
\end{algorithm}

\section{Computational Results}
\label{computationalresults}

To evaluate the performance of the Fast Temporal Pattern Mining with Extended Vertical Lists (FTPMwEVL) algorithm, we tested it against the approach by Batal \etal, from now on referred to as FTPM, on real-life datasets. The temporal pattern was defined as in Definition \ref{def:TP} for both algorithms.

All computations were carried out on a virtual server machine with 100 GB of memory and 20 virtual cores with processor speed equivalent to 2.5 GHz each. Only one core was utilized for single-thread computations. C++11 was used as a programming language. All computation times show actual pattern mining time taken by the algorithms after any preprocessing steps such as loading data and converting it into the abstraction domain.

It is important to state that the returned temporal patterns were entirely identical for both algorithms. It leaves computational time and memory usage as the only criteria for algorithm comparison.

\subsection{Acute Kidney Injury dataset}
The AKI dataset consists of $n = 5202$ medical records composed of time series taken during surgical procedures \cite{korenkevych2016pattern,thottakkara2016application}.
Each record has an outcome associated with it: 1, if Acute Kidney Injury (AKI) was diagnosed after the surgery (2769 records), and 0, otherwise (2433 records).

Using the University of Florida Integrated Data Repository,
we have previously assembled a single center cohort of perioperative
patients by integrating multiple existing clinical and administrative databases at
UF Health \cite{korenkevych2016pattern,thottakkara2016application}.
We included all patients admitted to the hospital for longer than 24 hours
 following any operative procedure between January 1, 2000, and November 30, 2010.
This dataset was integrated with the laboratory,
 the pharmacy and the blood bank databases and intraoperative 
database (Centricity Perioperative Management and Anesthesia, 
General Electric Healthcare, Inc.) to create a comprehensive intraoperative database for this cohort.
The study was designed and approved by the Institutional
 Review Board of the University of Florida and the University of Florida Privacy Office.

Two time variables were chosen for examination: Mean 
Arterial Blood Pressure (\textit{BP}) and Heart Rate (\textit{HR}).
The value abstractions were used to convert time series
 from time domain to abstraction domain
with percentile values $[0.1, 0.25, 0.75, 0.9]$ and support 
threshold $\theta$ (see Definition \ref{def:FTP}) ranging from $0.5$ to $0.9$.
The comparative data (see Table \ref{table:AKI}) indicates the superior performance of FTPMwEVL from the computational time point of view.
For $\theta = 0.7$, FTPMwEVL found all FTPs (there were no FTPs of size more than 18) in $39.58$ seconds using $3134.2$ megabytes of memory, while FTPM spent $34280.4$ seconds and $402.31$ megabytes to achieve the same result. Therefore, the speed-up was of magnitude  $\mathbf{866}$ while the new algorithm used only 7.79 times more memory. We set a computational time limit to 24 hours (86,400 seconds). In this time frame, FTPM was able to mine all FTPs only for $\theta \geq 0.7$. For $\theta =0.6$, FTPMwEVL found all FTPs 
(no FTPS of size more than 22), yet FTPM managed to mine FTPs of size 10 or lower and some of size 11. In this case, FTPMwEVL took $50.25$ seconds (not shown in the table) to mine all FTPs of size 11 or lower, and the speed-up column reflects ratio $86400~\text{sec} / 50.25~\text{sec} = 1719.3$. For $\theta = 0.5$, we limited the maximum FTP size to 12 due to FTPMwEVL memory consumption considerations. Still, FTPM mined only FTPs of size 7 or lower and some of size 8 in 86400 seconds. 

\begin{figure}[h]
\centering
\includegraphics[width=0.7\textwidth,trim={55 130 60 130},clip]{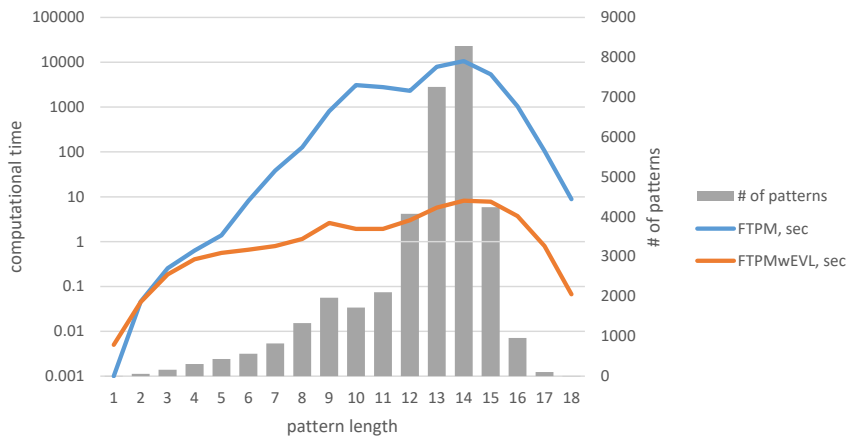}
\caption{%
Computational time in seconds for FTPMwEVL and FTPM on AKI dataset
for mining FTPs of sizes from 1 to 18 (there were no FTPs of a size larger than 19),
given that all FTPs of smaller sizes were already found.
Threshold $\theta$ was set at $0.7$. Total computational time was
$34280.4$ and $39.58$ seconds for FTPM and FTPMwEVL, respectfully.
Memory usage was $402.31$ and $3134.2$ megabytes.
Thus, FTPMwEVL achieved a significant speed-up of magnitude $\mathbf{866}$
while consuming $7.79$ times more memory.}
\label{fig:AKIandRand}
\end{figure}

As can be seen from Figure \ref{fig:AKIandRand}, the Extended Vertical Lists start working significantly better than the regular vertical lists with increasing FTP size, which happens due to the better indexing strategy that allows eliminating more candidate TPs and validating that a TP is not an FTP faster. Table \ref{table:AKI} demonstrates a phenomenon of exponential growth of computational time and memory usage with decreasing threshold level which is the main limitation of this pattern mining paradigm.

\begin{table*}
\caption{Computational time comparison of FTPMwEVL and FTPM on AKI datasets.}
\label{table:AKI}
\footnotesize
\centering
\begin{tabular}{|p{3mm}|p{6mm}|p{2mm}|p{12mm}|p{6mm}|p{2mm}|p{8mm}|p{10mm}|p{15mm}|p{10mm}|}\hline%
	\multirow{2}{*}{$\theta$}		&	\multirow{2}{4mm}{max $k$}	& \multicolumn{3}{c|}{FTPM} 				& \multicolumn{3}{c|}{FTPMwEVL}		&	\multirow{2}{15mm}{speed-up}			& \multirow{2}{5mm}{mem. ratio}\\\cline{3-5} \cline{6-8}
		&		&	k	&	sec				&	MB 			&	k	&	sec		&	MB			&				&	\\\hline\hline
0.9	&	Inf	&	7	&	0.16	&	1.86	&	7	&	0.1	&	15.8	&	1.06	&	8.49	\\\hline
0.8	&	Inf	&	12	&	465.5	&	24.92	&	12	&	2.3	&	198.5	&	$\mathbf{204.27}$	&	7.96	\\\hline
0.7	&	Inf	&	18	&	34280.4	&	402.31	&	18	&	39.6	&	3134.2	&	$\mathbf{866}$	&	7.79	\\\hline
0.6	&	Inf	&	10	&	$\mathbf{>86400}$	&	NA	&	22	&	621.1	&	35566.1	&	$\mathbf{>1719.3}$	&	NA	\\\hline
0.5	&	12	&	7	&	$\mathbf{>86400}$	&	NA	&	12	&	467.8	&	28950.4	&	$\mathbf{>998.93}$	&	NA	\\\hline
\end{tabular}
\end{table*}

\subsection{UCR Time Series Classification Archive}
The remaining datasets were taken from UCR Time Series Classification Archive \cite{UCRArchive}. Out of 85 datasets available, only those that have two classes were picked what resulted in 31 datasets. In this archive, each record has only one time series which was converted into two series of time-interval states using both trend and value abstractions.
Percentiles $[0.1, 0.25, 0.75, 0.9]$ for value abstractions were used to mine patterns in the UCR datasets, where, e.g., all values falling between percentiles 0.1 and 0.25 were considered as low. For trend abstractions, a segment was considered increasing if the slope was positive, and non-increasing, otherwise. The support threshold $\theta$ and maximum size $k$ varied in ranges $[0.2,0.8]$ and $[5,\infty]$, respectfully,  depending on the dataset complexity: we pushed the memory consumption of  FTPMwEVL to the limit. Therefore, Table \ref{table:UCR} reflects the most difficult cases from FTPMwEVL memory usage point of view.

FTPMwEVL was slower only on three datasets. The most significant speed-up of magnitude 3685.85 was achieved on dataset ``Computers''. For this case,  FTPMwEVL found all FTPs up to the predefined maximum size $k = 14$ (it was set on this level due to memory considerations) in 446.08 seconds having used 26100.1 megabytes of memory. FTPM mined all FTPs of size 8 or lower and some of size 9 in the time frame of 86400 seconds. Similar to the AKI dataset, the speed-up column reflects ratio $86400~\text{sec}/~23.44~\text{sec} = 3685.85$, where 23.44 seconds is the running time of FTPMwEVL to find all FTPs of size 9 or lower. Speed-up of 30 times or more was achieved on four other datasets: the values in bold font. However, after removing these outliers, the speed-up was on the level of $2.34$ on average for the remaining datasets.  The memory consumption was 4.15 time higher for FTPMwEVL on average. In the worst case, $35566.5$ megabytes of memory was allocated to store all FTPs which is not a concern for modern computational clusters.

\begin{table*}[h!]
\caption{Computational time comparison of FTPMwEVL and FTPM on UCR datasets.}
\label{table:UCR}
\scriptsize
\centering
\begin{tabular}{|p{19mm}|p{3mm}|p{4mm}|p{2mm}|p{12mm}|p{8mm}|p{2mm}|p{5mm}|p{8mm}|p{12mm}|p{5mm}|}\hline%
	\multirow{2}{*}{dataset}			&\multirow{2}{*}{$\theta$}		&	\multirow{2}{4mm}{max $k$}	& \multicolumn{3}{c|}{FTPM} 				& \multicolumn{3}{c|}{FTPMwEVL}		&	\multirow{2}{12mm}{speed-up}			& \multirow{2}{5mm}{mem. ratio}\\\cline{4-6} \cline{7-9}
									&		&		&	k	&	sec			&	MB 			&	k	&	sec		&	MB			&				&	\\\hline\hline
BeetleFly	&	0.8	&	8	&	8	&	702.0	&	5577.2	&	8	&	275.5	&	18367.7	&	2.55	&	3.29	\\\hline
BirdChicken	&	0.7	&	Inf	&	18	&	742.9	&	1564.4	&	18	&	269.9	&	4873.5	&	2.75	&	3.12	\\\hline
Coffee	&	0.8	&	10	&	10	&	560.1	&	4108.8	&	10	&	283.5	&	13107.4	&	1.98	&	3.19	\\\hline
Computers	&	0.8	&	14	&	8	&	$\mathbf{>86400}$	&	NA	&	14	&	446.1	&	26100.1	&	$\mathbf{>3685.8}$	&	NA	\\\hline
DistalPhalanx OutlineCorrect	&	0.7	&	Inf	&	16	&	160.3	&	1435.1	&	16	&	113.3	&	5382.6	&	1.41	&	3.75	\\\hline
Earthquakes	&	0.8	&	7	&	7	&	905.4	&	923.7	&	7	&	220.1	&	14340.2	&	4.11	&	15.52	\\\hline
ECG200	&	0.6	&	Inf	&	18	&	2349.7	&	3225.3	&	18	&	300.5	&	11863.1	&	7.82	&	3.68	\\\hline
ECGFiveDays	&	0.5	&	Inf	&	19	&	336.3	&	1181.4	&	19	&	226.1	&	3509.4	&	1.49	&	2.97	\\\hline
FordA	&	0.8	&	5	&	5	&	214.9	&	1462.2	&	5	&	113.4	&	11068.7	&	1.90	&	7.57	\\\hline
FordB	&	0.8	&	5	&	5	&	125.9	&	891.0	&	5	&	59.9	&	6461.7	&	2.10	&	7.25	\\\hline
Gun\_Point	&	0.2	&	Inf	&	18	&	35.3	&	129.6	&	18	&	27.0	&	377.7	&	1.31	&	2.91	\\\hline
Ham	&	0.8	&	7	&	7	&	783.8	&	5894.6	&	7	&	310.5	&	26318.9	&	2.52	&	4.46	\\\hline
HandOutlines	&	0.8	&	12	&	12	&	87.3	&	2847.4	&	12	&	134.9	&	9917.5	&	0.65	&	3.48	\\\hline
Herring	&	0.8	&	10	&	10	&	470.7	&	3186.7	&	10	&	177.9	&	10216.7	&	2.65	&	3.21	\\\hline
ItalyPowerDemand	&	0.2	&	Inf	&	13	&	1.2	&	16.8	&	13	&	1.4	&	46.4	&	0.86	&	2.75	\\\hline
Lighting2	&	0.8	&	8	&	8	&	50929.9	&	6080.6	&	8	&	495.1	&	35555.4	&	$\mathbf{102.87}$	&	5.85	\\\hline
MiddlePhalanx OutlineCorrect	&	0.7	&	Inf	&	17	&	301.6	&	2689.2	&	17	&	184.3	&	9633.6	&	1.64	&	3.58	\\\hline
MoteStrain	&	0.2	&	Inf	&	20	&	402.5	&	1109.5	&	20	&	380.6	&	2316.5	&	1.06	&	2.09	\\\hline
PhalangesOutlines Correct	&	0.5	&	Inf	&	14	&	96.6	&	1115.3	&	14	&	71.9	&	4035.1	&	1.34	&	3.62	\\\hline
ProximalPhalanx OutlineCorrect	&	0.4	&	Inf	&	19	&	588.0	&	4271.8	&	19	&	307.5	&	15383.3	&	1.91	&	3.60	\\\hline
ShapeletSim	&	0.8	&	7	&	6	&	$\mathbf{>86400}$	&	NA	&	7	&	378.1	&	25489.8	&	$\mathbf{>228.52}$	&	NA	\\\hline
SonyAIBORobot Surface	&	0.8	&	10	&	10	&	1387.0	&	5255.8	&	10	&	474.6	&	14650.9	&	2.92	&	2.79	\\\hline
SonyAIBORobot SurfaceII	&	0.8	&	10	&	10	&	2222.3	&	6619.7	&	10	&	701.0	&	21730.3	&	3.17	&	3.28	\\\hline
Strawberry	&	0.7	&	Inf	&	18	&	200.4	&	1701.7	&	18	&	105.9	&	6042.2	&	1.89	&	3.55	\\\hline
ToeSegmenta-tion1	&	0.8	&	8	&	8	&	754.7	&	2904.8	&	8	&	181.5	&	11128.0	&	4.16	&	3.83	\\\hline
ToeSegmenta-tion2	&	0.8	&	8	&	8	&	845.9	&	3076.6	&	8	&	188.2	&	10987.0	&	4.49	&	3.57	\\\hline
TwoLeadECG	&	0.2	&	Inf	&	17	&	15.1	&	93.5	&	17	&	15.2	&	243.1	&	0.99	&	2.60	\\\hline
wafer	&	0.7	&	Inf	&	11	&	$\mathbf{>86400}$	&	NA	&	29	&	275.6	&	10952.4	&	$\mathbf{>660.73}$	&	NA	\\\hline
Wine	&	0.7	&	Inf	&	22	&	481.8	&	4288.6	&	22	&	354.8	&	14043.6	&	1.36	&	3.27	\\\hline
WormsTwoClass	&	0.8	&	7	&	7	&	5459.6	&	2183.7	&	7	&	148.7	&	11036.3	&	$\mathbf{36.75}$	&	5.05	\\\hline
yoga	&	0.4	&	Inf	&	16	&	410.3	&	3593.7	&	16	&	215.6	&	8825.3	&	1.90	&	2.46	\\\hline
\end{tabular}
\end{table*}

\section{Concluding Remarks}
\label{conclusions}

In this paper, a new algorithm for Mining Frequent Temporal Patterns 
was presented where the concept of Extended Vertical List was utilized.
It outperformed the existing approach on many real-life datasets
in terms of computational time with minor exceptions. 
EVL requires more memory to be stored, which is a typical trade-off in such type of algorithms.
The proof of concept is that server clusters and personal computers 
have enough memory nowadays. Moreover, memory is becoming 
cheaper significantly faster than CPU, as well as the memory size
becomes five times of its previous size every two years (see http://www.jcmit.com/)
while CPU resources only double during the same time frame \cite{moore1975progress}.
Thus, the problem of using large amounts of memory is becoming less and less critical.

The speed-up was achieved due to EVL which works in two main directions:
 elimination of more candidate TPs, and faster verification of whether a candidate TP is an FTP or not. The candidate elimination by EVL works under the assumption that if a pattern is an FTP than all its subpatterns are FTPs as well. For other concepts of TP like Recent Temporal Pattern (RTP) in \cite{batal2016efficient}, this assumption does not hold. Thus, the candidate elimination phase will not work here, and only less efficient techniques like the Vertical Data Format should be utilized instead. Still, the concept of positions and indices will work for RTPs since the parent of an RTP is an RTP itself. Therefore, Extended Vertical Lists can give a partial speed-up for Frequent RTP Mining too. The approach can be generalized and applied to other domains where the notion of pattern is defined in other ways.

\section*{Acknowledgments}
AK was supported by the grant by University of Florida Informatics Institute. AB, PP and PM were supported by grant R01 GM-110240 by the National Institute of General Medical Sciences - National Institutes of Health. The content is solely the responsibility of the authors and does not necessarily represent the official views of the National Institutes of Health.

%% The Appendices part is started with the command \appendix;
%% appendix sections are then done as normal sections
%% \appendix

%% \section{}
%% \label{}

%\bibliographystyle{unsrt}
%\bibliography{lib}

\end{document}